# SDNIA-YOLO: A Robust Object Detection Model for Extreme Weather Conditions

Yuexiong Ding, Xiaowei Luo

*Abstract*—Though current object detection models based on deep learning have achieved excellent results on many conventional benchmark datasets, their performance will dramatically decline on real-world images taken under extreme conditions. Existing methods either used image augmentation based on traditional image processing algorithms or applied customized and scene-limited image adaptation technologies for robust modeling. This study thus proposes a stylization data-driven neural-image-adaptive YOLO (SDNIA-YOLO), which improves the model's robustness by enhancing image quality adaptively and learning valuable information related to extreme weather conditions from images synthesized by neural style transfer (NST). Experiments show that the developed SDNIA-YOLOv3 achieves significant mAP@.5 improvements of at least 15% on the real-world foggy (RTTS) and lowlight (ExDark) test sets compared with the baseline model. Besides, the experiments also highlight the outstanding potential of stylization data in simulating extreme weather conditions. The developed SDNIA-YOLO remains excellent characteristics of the native YOLO to a great extent, such as end-to-end one-stage, data-driven, and fast.

## I. INTRODUCTION

Object detection is one of the essential tasks of computer vision. Currently, deep learning methods based on the convolutional neural network (CNN) have become mainstream backbones for object detection and have achieved excellent performance on many conventional benchmark datasets [1]–[4]. However, object detection models trained on conventional images usually lacked robustness, achieving unsatisfactory results in extreme weather conditions, such as foggy and lowlight conditions [5].

There are many relevant studies on de-raining [6]–[8], de-hazing [9]–[11], and lowlight enhancement [12]. However, it is impractical to directly combine these models with the object detection network since they are deep and complex CNN models and need to be trained separately. These additional deep CNN models would make the final integrated model unable to meet the most basic real-time requirements of object detection. On the other hand, native YOLO models have used various traditional algorithms (TAs) of image processing to simulate extreme conditions for the input images [1], [13], such as random noise, random gamma transform, random blur, and random brightness and contrast [14]. However, the effect of TAs is limited since those TAs-enhanced models still have sharp performance declines when encountering extreme weather conditions.

The latest relevant research on object detection in adverse weather conditions is IA-YOLO [5], which proposed a white-box image-adaptive (IA) module to enhance images before input into the YOLOv3 to obtain better detection results. However, the white-box IA is scene-dependent, which means that the IA-YOLO needs to customize the designs of different IA modules and re-train them for different extreme scenes. Besides, the white-box strategy is unnecessary in practical applications since most of them only care about the final detection results and do not focus on which processing operation the image needs to be executed.

This study proposed a stylization data-driven neural-image-adaptive YOLO (SDNIA-YOLO) for robust object detection in extreme weather conditions. The SDNIA-YOLO comprises a neural-image-adaptive (NIA) module and a YOLO backbone. The NIA module is a lightweight CNN model for image adaptation, which learns to adaptively restore images to their optimal state by eliminating information related to extreme weather conditions. The adjusted images are then input into the YOLO to complete the detection task. Instead of designing a series of adaptive algorithms for each scene alone, the proposed NIA module is data-driven that can learn corresponding adaptation abilities for specific extreme conditions as long as the training data contains specific extreme scenes. The neural style transfer (NST) technology is used in this study for extreme conditions simulation, which endows training data with specific extreme information by transferring extreme conditions from style images to conventional/normal training data. The NIA module and the YOLO backbone are jointly trained end-to-end using stylized and conventional images. Finally, the developed SDNIA-

The Shenzhen Science and Technology Innovation Committee Grant #JCYJ201805071781647320 and the General Research Fund from the Research Grant Council of Hong Kong SAR #11211622 jointly supported this work. The conclusions herein are those of the authors and do not necessarily reflect the views of the sponsoring agencies. *(Corresponding author: Xiaowei Luo).*

Yuexiong Ding is with the Department of Architecture and Civil Engineering, City University of Hong Kong, Hong Kong, China, and Architecture and Civil Engineering Research Center, Shenzhen Research Institute of City University of Hong Kong, Shenzhen, China.

Jia Pan is with the Department of Computer Science, University of Hong Kong, Hong Kong, China.

Jinxing Hu is with the Shenzhen Institute of Advanced Technology, Chinese Academy of Sciences, Shenzhen 518055, China.

Xiaowei Luo is with the Department of Architecture and Civil Engineering, City University of Hong Kong, Hong Kong, China, and Architecture and Civil Engineering Research Center, Shenzhen Research Institute of City University of Hong Kong, Shenzhen, China (e-mail: xiaowluo@ cityu.edu.hk).

YOLO outperforms the baseline YOLO and existing IA-YOLO and achieves impressive mAP improvements in foggy (RTTS) and lowlight (ExDark) scenarios, as shown in **Fig. 1**.

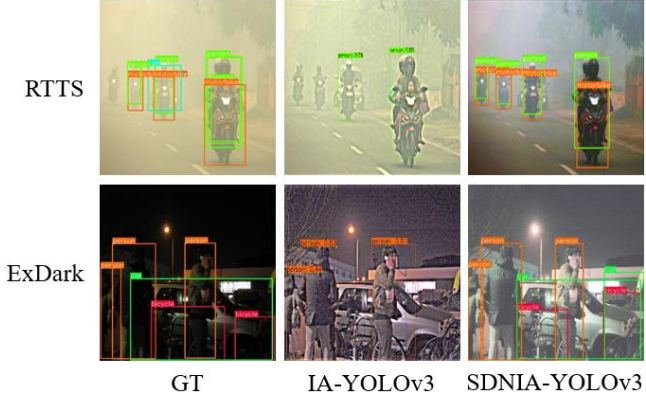

**Fig. 1.** Performance comparison between the IA-YOLOv3 and SDBIA-YOLOv3.

The contributions of this study can be summarized as follows: 1) This study proposes a new SDNIA-YOLO model for robust object detection in extreme weather conditions, which is further verified to achieve significant mAP improvements on the real-world foggy and lowlight test sets; 2) The proposed NIA is a lightweight data-driven module, making the developed SDNIA-YOLO inherit native YOLO's many excellent features to a great extent, such as end-to-end/one-stage, scene-independent, and fast inference; 3) This study also verifies the feasibility and application value of using the NST technology for extreme weather conditions simulation in improving model robustness.

The rest of the paper is organized as follows: Section 2 reviews some related works to provide research foundations. Section 3 describes the proposed SDNIA-YOLO in detail. Section 4 conducts experiments on foggy and lowlight scenarios, while Section 5 further discusses and analyzes the performance of the SDNIA-YOLO. Finally, Section 6 concludes this work.

## II. Related works

**Object detection.** The current mainstream CNN-based object detection methods can be divided into two categories. One is the two-stage Region-CNNs (R-CNNs) based on the regional proposals, which first generate regions of interest (ROI) from images and then classify them by another neural network [15]–[17]. The other is based on one-stage regression, like the YOLOs [1]–[3], [18] and single shot multibox detector (SSD) [19], which predict objects labels and bounding boxes coordinates at once using only one regression network. In this study, the proposed model applied the classic YOLO architecture as the model backbone, not only because of its one-step, fast and high-precision features but also for direct comparison with the latest similar research IA-YOLO [5].

**Object detection in extreme conditions.** Currently, there are mainly three kinds of solutions for object detection in extreme conditions. The most common one is the image augmentation method that applies various transformation functions (e.g., random blur, random rain, random gamma, random brightness, etc.) with a certain probability to simulate extreme conditions [13], [14]. The second is to perform image enhancement and detection tasks simultaneously through joint or multi-task learning [5], [20], [21]. However, these methods design loss functions based on specific physical/empirical formulas. The last one is domain-adaptive [22]–[24], which assumes that there is domain transfer between images captured under normal and poor weather conditions, and the weather-specific information can be eliminated by learning the domain prior knowledge to make features weather-invariant.

**Image adaptation (IA).** Image adaptation is an emerging technology of image enhancement. Traditional methods generally calculate the hyperparameters of the designed transformation functions based on image features according to some empirical formulas [25]–[27]. For example, Wang et al. [26] applied local gamma transform and color compensation to adaptively adjust the enhancement parameters based on the illumination distribution characteristics of the input image. With the development of deep learning, CNNs are started to be increasingly used to learn hyperparameters automatically for various transformation functions according to the extracted image features [5], [28]–[30]. For example, to obtain better detection results in extreme weather conditions, Liu et al. developed a white-box IA module consisting of a series of image transformation algorithms to enhance images before input into the object detection model, whose parameters were predicted by a small CNN simultaneously and adaptively. However, this kind of white-box adaptive method must know in advance how many transformation functions and which hyperparameters need to be learned. Besides, the design of the white-box IA module needs to be customized and re-trained for specific extreme conditions.

**Neural style transfer (NST).** The NST was proposed to transfer styles from style images to content images with minimum content loss but maximum style similarity [31]. In 2017, Ghiasi et al. [32] proposed an arbitrary NST, which performed style transfer on any given content and style image using only one model. The NST can naturally create randomness at the image level since the content of each image is different. Besides, the Gram matrix used in style loss of the NST model paid attention to style texture and ignored global spatial information, increasing the style randomness at the pixel level in an image after stylizations. Owning such advantages, the NST has been applied to image classification tasks as a new augmentation method [33]–[35], but no attempt has been explored in object detection and extreme condition simulation.

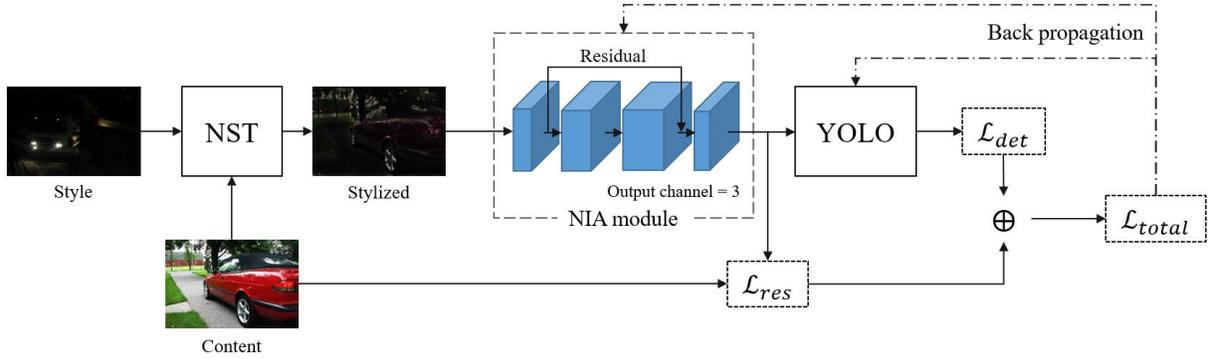

**Fig. 2.** The framework of the SDNIA-YOLO.

## III. METHODOLOGY

**Fig. 2** is the overall framework of the proposed SDNIA-YOLO. The stylized images are first synthesized via an NST model to simulate extreme weather conditions. These synthesized data are then mixed with the original content images to form the training data. In model training, two data pipelines are produced after the NIA module: one is used to calculate the image restoration loss ($\mathcal{L}_{res}$) together with the original content images, the other input directly into the YOLO model to calculate the detection loss ($\mathcal{L}_{det}$). Finally, the parameters of the NIA and YOLO are optimized jointly in each training batch according to the weighted loss of $\mathcal{L}_{res}$ and $\mathcal{L}_{det}$.

*A. Neural style transfer*

Embedding real-world extreme conditions into conventional images is promising for simulating extreme conditions. An excellent embedding method should transfer the extreme scene style to the image without changing its essential content, and the NST model is quite suitable for achieving such functions. **Fig. 3** shows the stylization procedures of the arbitrary NST model: a style prediction network ($P$) is first used to predict the style vector ($\vec{S}$) from the style image; the content image is then stylized by the style transfer network ($T$) under the constrain of the predicted style vector $\vec{S}$. Such a modular structure design allows easy controlling of the stylization strength ($\alpha$) by using the "identity interpolation" [32]. **Fig. 4** demonstrates some stylization images using different strengths. As $\alpha$ increases, more details of image content is losing, and the embedded style is gradually closing to extreme conditions. The NST model can generate multiple images with various extreme conditions for one conventional content image using diverse style images with different stylization strengths. In theory, the total number of synthetic images can be calculated by $N_c \times N_s \times N_\alpha$, where $N_c$, $N_s$, and $N_\alpha$ are the number of content images, style images, and strength factors, respectively.

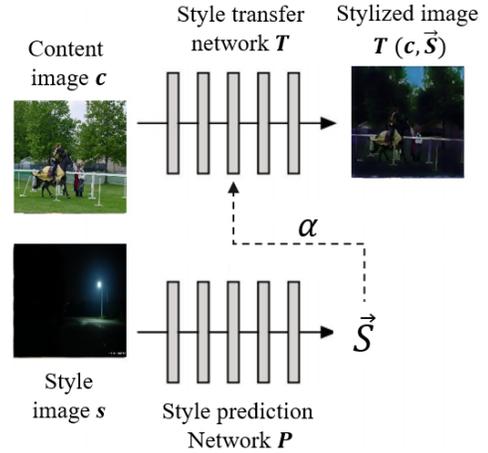

**Fig. 3.** Stylization procedures of the arbitrary NST model.

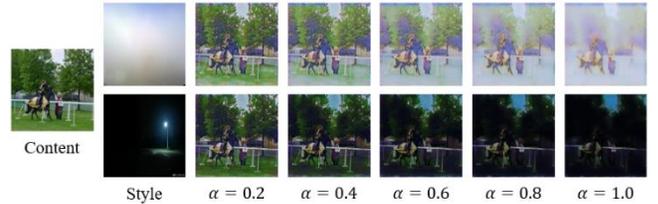

**Fig. 4.** Stylization using different strengths ($\alpha$).

*B. Neural-image-adaptive module*

The NIA module primarily aims to remove the information related to extreme conditions from images before conducting object detection. Unlike the white-box IA module, the NIA is a black-box module focusing only on the final restoration results, while operations and transformations executed inside the module are opaque. Considering the inference efficiency of the final end-to-end model, the NIA module is designed as a lightweight CNN, as shown in **Fig. 5**. The output of each inner NIA layer is set to the same size since the network is relatively shallow and unsuitable for applying the encoding-decoding strategy. Like other ordinary CNNs, the NIA is also data-driven, which means that its ability to eliminate extreme information depends on what kinds of training data it has been fed.

| Module | Layer | Filters | size | Repeat | Output size |
|---|---|---|---|---|---|
|  | Image |  |  |  | 544×544 |
|  | Conv | 32 | 3×3 / 1 | 1 |  |
|  | Conv | 64 | 3×3 / 1 | 1 |  |
| NIA | Conv | 32 | 1×1 / 1 | Conv | 544×544 |
|  | Conv | 64 | 3×3 / 1 | Conv ×1 |  |
|  | Residual |  |  | Residual |  |
|  | Conv | 3 | 3×3 / 1 | 1 |  |
| YOLOv3 | Conv | 32 | 3×3 / 1 | 1 | 544×544 |
|  | Conv | 64 | 3×3 / 2 | 1 | 272×272 |
|  | ... |  |  |  |  |

**Fig. 5.** The network architecture of the NIA module.

Regarding image content restoration loss, the $l1$ or $l2$ loss are intuitively preferred. However, these pixel-wise methods only compare differences pixel by pixel without considering human visual perception and aesthetics. Minimum $l1/l2$ loss does not necessarily mean good image restoration. Therefore, the multiscale structural similarity index measure (MS-SSIM) combined with the $l1$ loss is used to consider human visual perception, including brightness, contrast, and structure in different image resolution levels [36]. However, the MS-SSIM + $l1$ loss still not consider inherent or high-level features restoration of the image. The perceptual loss based on the VGG model [37] thus is adopted to take the high-level features restoration loss into account [38]. Finally, the image restoration loss consists of the $l1$, MS-SSIM, and VGG-based perceptual loss, as represented in (1), where $\alpha, \beta,$ and $\gamma$ were set to 0.25, 0.25, and 0.5 empirically in this study during training.

$$\mathcal{L}_{res} = \alpha \cdot \mathcal{L}_{l1} + \beta \cdot \mathcal{L}_{MS-SSIM} + \gamma \cdot \left( \mathcal{L}_{content}^{VGG} + \mathcal{L}_{style}^{VGG} \right) \quad (1)$$

*C. Object detection backbone*

Considering the excellent performance of the YOLO families in the object detection area, the YOLO architecture was adopted as the object detection backbone in this study. Specifically, the proposed SDNIA-YOLO was modified from the open-source YOLO (v3 and v5) implementation [14]. The object detection loss includes bounding box loss, confidence loss, and classification loss, as shown in (2), where $p_1$, $p_2$ and $p_3$ were set to 0.05, 1.0 and 0.5 empirically in this study during training.

$$\mathcal{L}_{det} = p_1 \cdot \mathcal{L}_{box} + p_2 \cdot \mathcal{L}_{obj} + p_3 \cdot \mathcal{L}_{cls} \quad (2)$$

*D. Model training and inference*

The model training of the SDNIA-YOLO is slightly different from the native YOLO, mainly lying in model input and output. As shown in **Fig. 2**, the input of the SDNIA-YOLO in the training stage included images to be detected and images for restoration reference. The reference image is the original content image without any stylization. During model training, the NIA and YOLO are optimized using the joint/multi-task learning strategy in each training batch according to the total loss, as shown in (3), where $p_4$ was empirically set to 0.01 in this study. In the inference phase, the reference original image is not needed by the SDNIA-YOLO model, which becomes consistent with the native YOLO, only requiring the image to be detected as the input.

$$\mathcal{L}_{total} = p_1 \cdot \mathcal{L}_{box} + p_2 \cdot \mathcal{L}_{obj} + p_3 \cdot \mathcal{L}_{cls} + p_4 \cdot \mathcal{L}_{res} \quad (3)$$

## IV. EXPERIMENTS

*A. Data preparation*

Consistent with research [5], two extreme scenarios were considered in this study to conduct the experiments. Images from the VOC2007_trainval and VOC2012_trainval sets [39] containing **Person**, **Bicycle**, **Car**, **Bus**, or **Motorcycle** objects were extracted to construct the new training&validation (VOC_trainval) set for the foggy scenario, while images containing **Person** (People), **Bicycle**, **Car**, **Bus**, **Motorcycle** (Motorbike), **Boat**, **Bottle**, **Cat**, **Chair**, or **Dog** were extracted to construct the VOC_trainval set for the lowlight scenario. The extracted images were then stylized ($\alpha = 1.0$) via a well-trained arbitrary NST model [40] using the collected style images and mixed with the original images to form the final training&validation set named VOC_trainval_mixed (*VTM*). Thirteen style images for each extreme scene were randomly collected online for demonstration. The collection criteria of style images were relatively simple: focusing only on the required extreme style without considering the image content since only the style would be transferred.

As for model evaluation, three different test sets were collected for each extreme scene. First is the VOC_norm_test (*VNT*), extracted from the VOC2007_test set using the same construction method of the VOC_trainval. Then are two synthesized test sets from [5], VOC_foggy_test (*VFT*) and VOC_dark_test (*VDT*), which were synthesized based on the *VNT* set via the foggy equation and random gamma transformation. Finally are two real-world datasets, *RTTS* [41] in foggy conditions and ExDark_test (*EDT*) set [42] in lowlight conditions. TABLE I shows detailed information about each dataset, and Fig. **6** shows some extreme style images and the corresponding stylization results.

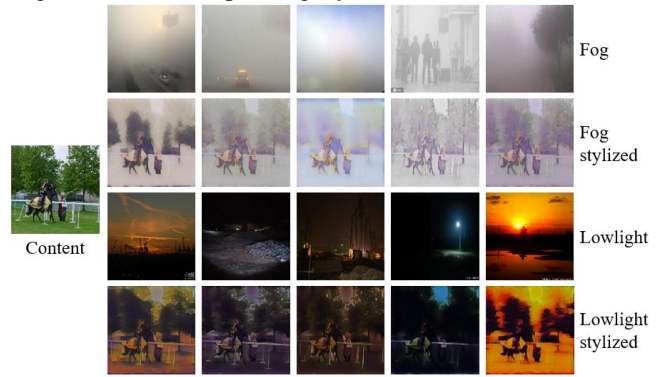

**Fig. 6.** Examples of extreme style images (the first and third rows) and the corresponding stylization ($\alpha = 1.0$) results (the second and fourth rows).

*B. Experiment settings*

The SDNIA-YOLO model was separately trained for the foggy and lowlight scenarios. During model training, some key parameters were set as follows: the optimizer was SGD with a learning rate of 0.001, the input batch size was four images of 544×544, and the training epoch was set to a large number (e.g., 400), but the early stopping strategy with the patience of ten was applied. After training, two MS COCO standard metrics of mean average precision (mAP) [43], mAP@.5 and mAP@.5: .95, were applied for evaluation, where mAP@.5 denotes the mAP calculated upon the 0.5 intersection-over-union (*IoU*) threshold and mAP@.5: .95 means the average mAP over different *IoU* thresholds from 0.5 to 0.95 with a step of 0.05. All experiments were conducted in the "*Python3.6 + RTX 2080 Ti*" environment.

TABLE I
DETAILS OF EACH DATASET USED IN DIFFERENT EXTREME SCENES

| Scene | Classes | Style Images | VTM ($\alpha = 1.0$) | VNT | VFT | VDT | RTTS | EDT |
|---|---|---|---|---|---|---|---|---|
| Foggy | 5 | 13 | (13 × 8,111) + 8,111 | 2,734 | 2,734 | / | 4,322 | / |
| Lowlight | 10 | 13 | (13 × 12,334) + 12,334 | 3,760 | / | 3,760 | / | 2,563 |

TABLE II
PERFORMANCE COMPARISON IN THE FOGGY SCENARIO

| Model | VNT | | VFT | | RTTS | |
|---|---|---|---|---|---|---|
| | mAP@.5 | mAP@.5: .95 | mAP@.5 | mAP@.5: .95 | mAP@.5 | mAP@.5: .95 |
| DAYOLO | 56.51 | / | 55.11 | / | 29.93 | / |
| DSNet | 53.29 | / | 67.40 | / | 28.91 | / |
| IA-YOLOv3 | 73.23 | / | 72.03 | / | 37.08 | / |
| YOLOv3 | 86.09 | 64.05 | 74.36 | 52.33 | 50.27 | 31.53 |
| YOLOv5 | 88.90 | 67.61 | 78.13 | 56.29 | 54.86 | 33.86 |
| SDNIA-YOLOv3 | 92.27 | 72.56 | 90.92 | 71.18 | 66.0 | 40.81 |
| SDNIA-YOLOv5 | 93.15 | 74.92 | 92.11 | 73.69 | 66.71 | 42.64 |

* No data from RTTS were used for model training.

TABLE III
PERFORMANCE COMPARISON IN THE LOWLIGHT SCENARIO

| Model | VNT | | VDT | | EDT | |
|---|---|---|---|---|---|---|
| | mAP@.5 | mAP@.5: .95 | mAP@.5 | mAP@.5: .95 | mAP@.5 | mAP@.5: .95 |
| DAYOLO | 41.68 | / | 21.53 | / | 18.15 | / |
| DSNet | 64.08 | / | 43.75 | / | 36.97 | / |
| IA-YOLOv3 | 70.02 | / | 59.40 | / | 40.37 | / |
| YOLOv3 | 80.78 | 58.49 | 67.65 | 45.16 | 57.38 | 30.07 |
| YOLOv5 | 83.06 | 61.19 | 69.69 | 48.87 | 59.83 | 33.18 |
| SDNIA-YOLOv3 | 87.33 | 67.72 | 82.19 | 60.5 | 72.5 | 41.94 |
| SDNIA-YOLOv5 | 89.24 | 71.58 | 83.5 | 64.49 | 75.25 | 43.92 |

* No data from ExDark were used for model training.

*C. Results*

1) **Performance comparison**

TABLE II and TABLE III compare the performance of the baseline YOLOv3/v5 and SDNIA-YOLOv3/v5 in the foggy and lowlight scenarios. Note that the YOLOv5x backbone was actually used in this study but is still labeled as YOLOv5 below for convenience. Besides, the performances of six relevant models/methods developed for handling poor environmental conditions are also listed in the two tables, which were the experimental results given by the IA-YOLOv3 research [5]. The DAYOLO [44] is a domain-adaptive model, and the DSNet [20] is a model based on multi-task learning.

First, the SDNIA-YOLO models perform best compared with others in foggy and lowlight scenarios. Notably, the SDNIA-YOLO models perform much better than the IA-YOLO on all test datasets, indicating that the developed black-box SDNIA module is better at handling extreme weather conditions than the white-box IA module. Besides, though the baseline YOLOv3 and YOLOv5 achieve satisfactory performance on VOC normal test data (*VNT*), their performances dramatically decline on the synthesized and real-world extreme test data. The YOLOv3 suffers 11.73%, 35.82%, 13.13%, and 23.4% mAP@.5 drops on the *VFT*,

*RTTS*, *VDT*, and *EDT* sets, while these data of the YOLOv5 are 10.77%, 34.04%, 13.37%, and 23.23%, respectively. The SDNIA-YOLO models significantly improve the situation: the SDNIA-YOLOv3 achieves 16.56%, 15.73%, 14.54%, and 15.12% mAP@.5 improvements on the *VFT*, *RTTS*, *VDT*, and *EDT* sets compared with the baseline YOLOv3, while the SDNIA-YOLOv5 achieves corresponding improvements of 13.98%, 11.85%, 13.81%, and 15.42% compared with its baseline model. In addition to being more robust to extreme conditions, the SDNIA-YOLO models also achieve higher performance in normal conditions. For example, the SDNIA-YOLOv3 models built for two extreme scenarios achieve 7.06% and 6.55% mAP@.5 improvements on the *VNT* set.

In short, the proposed black-box SDNIA module is better at handling extreme weather conditions than the existing white-box IA module, and the SDNIA-YOLOv3 achieves impressive mAP@.5 improvements of rough 15% on the real-world foggy and lowlight datasets compared with the baseline model.

2) **Results visualization**

Fig. **7** and **Fig. 8** show some examples of image adaptation and detection results of the IA-YOLOv3 and SDNIA-YOLOv3 in foggy and lowlight scenarios. From the perspective of the image adaptation effect, the IA module relatively over-sharpens the input images and generates many noises in the lowlight scenario, while the SDBIA module outputs softer images with more natural lighting. Nonetheless, it is still hard to determine which module achieves the best adaptation effect if relying only on human-eye observation. When comparing the detection results, the performance of the SDNIA-YOLOv3 is highlighted immediately. The IA-YOLOv3 un-detects many key objects and even detects nothing in some cases (e.g., images A4 and C4 in Fig. **7** and B6 and C6 in **Fig. 8**), while the SDNIA-YOLOv3 almost perfectly detects all considered objects and even detects correct objects that were unlabeled in the ground truth (e.g., images C5 in Fig. **7** and C5 in **Fig. 8**).

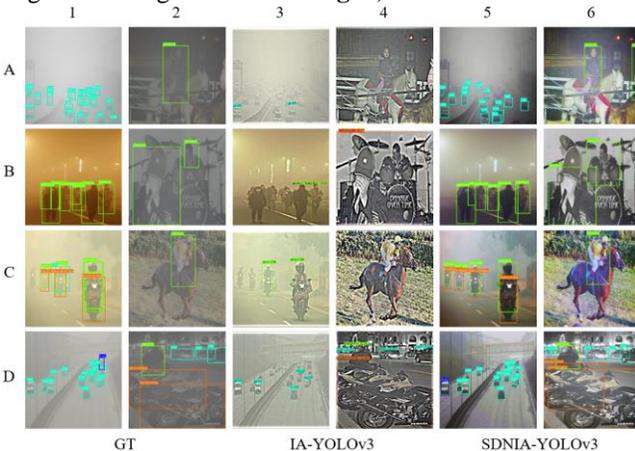

**Fig. 7.** Image adaptation and detection results of the IA-YOLOv3 and SDNIA-YOLOv3 in the foggy scenario. Columns 1, 3, and 5 are images from the *RTTS*; columns 2, 4, and 6 are images from the *VFT*.

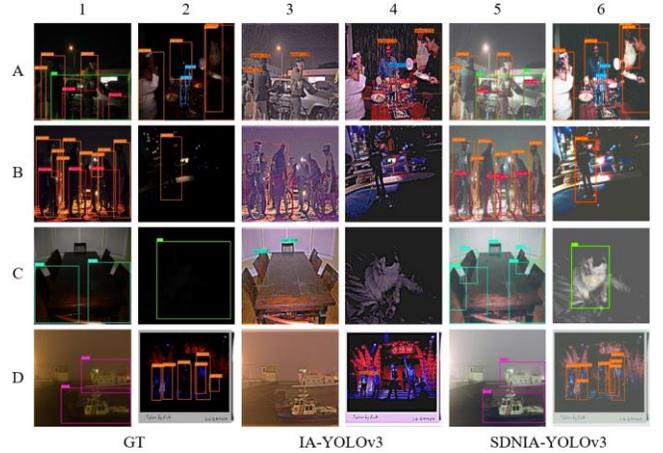

**Fig. 8.** Image adaptation and detection results of the IA-YOLOv3 and SDNIA-YOLOv3 in the lowlight scenario. Columns 1, 3, and 5 are images from the *EDT*; columns 2, 4, and 6 are images from the *VDT*.

3) **Efficiency analysis**

TABLE IV compares the model scale and inference speed of the baseline YOLOv3, IA-YOLOv3, and SDNIA-YOLOv3. The developed SDNIA-YOLOv3 introduces about 42K extra trainable parameters, only about 1/4 of the parameters introduced by IA-YOLOv3. As for the inference speed, the developed SDNIA-YOLOv3 saves nearly 25*ms* to detect a 544×544 image compared with the IA-YOLOv3 and only costs additional 5*ms* compared with the native YOLOv3. Even in processing high-resolution images of 1024 × 1024, the SDNIA-YOLOv3 can achieve a speed of about 20 frames per second, which still meets the essential requirement of real-time.

TABLE IV
EFFICIENCY ANALYSIS OF THE IA-YOLOV3 AND SDNIA-YOLOV3

| Model | Additional parameters | GPU | Speed (*ms*) 544×544 | 1024×1024 |
|---|---|---|---|---|
| YOLOv3 | +0 | RTX2080Ti | 14.4 | 34.7 |
| IA-YOLOv3 | +165K | TeslaV100 | 44 | / |
| SDNIA-YOLOv3 | +42K | RTX2080Ti | 19.3 | 49 |

## V. ABLATION STUDIES

In this section, several ablation experiments based on the SDNIA-YOLOv3 and the foggy scenario were conducted to explore the influence of different components on the model's performance.

*A. Impact of image restoration loss*

As shown in TABLE V, though the model with MS-SSIM+$l$1 loss achieves higher mAP@.5 on the *VNT* and *VFT* sets, the model with MS-SSIM+$l$1+ VGG_Perceptual loss avhieves more comprehensive performance with the highest

mAP@.5: .95 on all test sets. In addition, the model with the MS-SSIM+$l1$ loss performs better than the model with the VGG_P loss on all test sets, indicating that it might be better to restore human visual perceptible information than restore the inherent high-level features of the image when conducting object detection in extreme conditions. In other words, the current object detection model relies more on human visual perceptible information in the image than on its high-level features, providing indirect evidence of the necessity of conducting image adaptation before object detection.

TABLE V
THE IMPACT OF THE IMAGE RESTORATION LOSS. (MAP@.5 / MAP@.5: .95)

| M+$l1$ | VGG_P | VNT | VFT | RTTS |
|---|---|---|---|---|
| ✓ |  | **92.98** / 71.54 | **92.05** / 70.36 | **66.0** / 40.14 |
|  | ✓ | 90.79 / 71.41 | 89.25 / 69.79 | 61.61 / 38.7 |
| ✓ | ✓ | 92.27 / **72.56** | 90.92 / **71.18** | **66.0** / **40.81** |

*B. Impact of stylization data and the NIA module*

As shown in TABLE VI, applying either stylization data or the NIA module alone can help the baseline model achieve encouraging performance improvements, while applying them together achieves the best comprehensive performance with the highest mAP@.5: .95 on all test sets. Besides, the encouraging high performance of the SD-YOLOv3 and NIA-YOLOv3 on three test sets indicates that learning the knowledge of extreme conditions in advance and enhancing image quality are two efficient ways to make a more robust object detection model in poor weather conditions. Finally, the encouraging performance of the SD-YOLOv3 also reflects the potential superiorities of the NST technology in simulating extreme weather conditions.

TABLE VI
THE IMPACT OF THE NIA MODULE AND STYLIZATION DATA (MAP@.5 / MAP@.5: .95)

| Model | SD | NIA | VNT | VFT | RTTS |
|---|---|---|---|---|---|
| YOLOv3 |  |  | 88.09 / 64.05 | 76.36 / 52.33 | 54.27 / 32.53 |
| SD-YOLOv3 | ✓ |  | 90.09 / 69.22 | 87.63 / 66.84 | 60.57 / 37.31 |
| NIA-YOLOv3 |  | ✓ | 93.31 / 72.4 | 91.33 / 69.94 | 64.12 / 39.97 |
| SDNIA-YOLOv3 | ✓ | ✓ | 92.27 / 72.56 | 90.92 / 71.18 | 66.0 / 40.81 |

*C. Impact of stylization strength*

The combination of stylization data with different $\alpha$ also has a certain impact on the model's performance. Therefore, the SDNIA-YOLOv3 were trained on different stylization datasets with different $\alpha$ ranges from 0.2 to 1.0 with a step of 0.2, as shown in TABLE VII, where range [0.6: 1.0] means applying stylization with $\alpha$ values of [0.6, 0.8, 1.0]. As the $\alpha$ range increases, the performance of the SDNIA-YOLOv3 is gradually getting better and reaches the best between the $\alpha$ range of [0.8: 1.0] and [0.4: 1.0]. According to **Fig. 4**, there is little difference between the stylized images and the source content images when $\alpha$ is in a small value, which is equivalent to introducing repeated images without contributing new knowledge to the model, explaining why the performance stops growing after [0.4: 1.0]. Therefore, stylization data with relatively high $\alpha$ can provide the model with valuable information related to extreme weather conditions, and the combination of these $\alpha$ provides a more comprehensive knowledge domain for the model to achieve higher performance.

TABLE VII
THE IMPACT OF THE NIA MODULE AND STYLIZATION DATA (MAP@.5 / MAP@.5: .95)

| $\alpha$ range | VNT | VFT | RTTS |
|---|---|---|---|
| 1.0 | 92.27 / 72.56 | 90.92 / 71.18 | 66.0 / 40.81 |
| [0.8: 1.0] | 92.93 / **72.93** | 91.82 / **71.7** | 66.59 / 41.7 |
| [0.6: 1.0] | 93.62 / 72.11 | 92.77 / 71.08 | **68.93** / **42.2** |
| [0.4: 1.0] | **93.79** / 70.42 | **92.99** / 69.76 | 68.54 / 41.32 |
| [0.2: 1.0] | 93.28 / 70.36 | 92.24 / 69.11 | 68.54 / 40.59 |

VI. CONCLUSIONS

This study proposes the SDNIA-YOLO model for more robust object detection in extreme weather conditions. Experiments and ablation studies on foggy and lowlight scenarios verify the feasibility and efficiency of the proposed model. First, the SDNIA-YOLO outperforms the other models and achieves impressive mAP@.5 improvements of at least 15% on the real-world foggy and lowlight test sets. Besides, the NST technology has great potential in simulating poor weather conditions, and the stylization data can provide much valuable extreme conditions knowledge for the detection model. Driven by the stylization data, the CNN-based NIA module can eliminate extreme conditions-related information adaptively for the input images before conducting object detection. Future work can focus on validating models in more scenarios (e.g., rainy, snowy, overexposed, shaded, etc.) or implementing a plug-and-play SDNIA module to make it a generic tool for other computer vision tasks.